\documentclass[letterpaper, 10 pt, conference]{ieeeconf}  % Comment this line out if you need a4paper

\IEEEoverridecommandlockouts                              % This command is only needed if 
\usepackage[utf8]{inputenc}
\usepackage[T1]{fontenc}
\usepackage{amsfonts,amsmath,amssymb}
\usepackage{anyfontsize}
\usepackage{booktabs}
\usepackage{graphicx}
\usepackage{mathtools}
\usepackage{microtype}
\usepackage{multirow}
\usepackage{subcaption}
\usepackage[hyphens]{url}
\usepackage[table]{xcolor}
\usepackage{xspace}
\usepackage[pagebackref]{hyperref}
\hypersetup{%
	pdfauthor={Dongxu Guo, Taylor Mordan, Alexandre Alahi},%
	pdftitle={Pedestrian Stop and Go Forecasting with Hybrid Feature Fusion},%
	colorlinks=true,%
	linkcolor=blue,%
	citecolor=green%
}
\usepackage[nameinlink]{cleveref}

\title{\LARGE \bf
Pedestrian Stop and Go Forecasting with Hybrid Feature Fusion
}

\author{Dongxu Guo, Taylor Mordan and Alexandre Alahi% <-this % stops a space
\thanks{All authors are with École Polytechnique Fédérale de Lausanne (EPFL), Visual Intelligence for Transportation (VITA), CH-1015 Lausanne, Switzerland.}%
}

\begin{document}

\maketitle
\thispagestyle{empty}
\pagestyle{empty}

%%%%%%%%%%%%%%%%%%%%%%%%%%%%%%%%%%%%%%%%%%%%%%%%%%%%%%%%%%%%%%%%%%%%%%%%%%%%%%%%
\begin{abstract}
Forecasting pedestrians' future motions is essential for autonomous driving systems to safely navigate in urban areas. However, existing prediction algorithms often overly rely on past observed trajectories and tend to fail around abrupt dynamic changes, such as when pedestrians suddenly start or stop walking. We suggest that predicting these highly non-linear transitions should form a core component to improve the robustness of motion prediction algorithms. In this paper, we introduce the new task of pedestrian stop and go forecasting. Considering the lack of suitable existing datasets for it, we release TRANS, a benchmark for explicitly studying the stop and go behaviors of pedestrians in urban traffic. We build it from several existing datasets annotated with pedestrians' walking motions, in order to have various scenarios and behaviors. We also propose a novel hybrid model that leverages pedestrian-specific and scene features from several modalities, both video sequences and high-level attributes, and gradually fuses them to integrate multiple levels of context. We evaluate our model and several baselines on TRANS, and set a new benchmark for the community to work on pedestrian stop and go forecasting.
\end{abstract}

%%%%%%%%%%%%%%%%%%%%%%%%%%%%%%%%%%%%%%%%%%%%%%%%%%%%%%%%%%%%%%%%%%%%%%%%%%%%%%%%
\section{Introduction}

When navigating in populated cities, autonomous vehicles need to anticipate the future movements of pedestrians, who are arguably among the most vulnerable road users, and react accordingly to avoid potential collisions \cite{alahi2008object,ridel2018a}.
A large body of work on pedestrian motion prediction uses past observed trajectories to forecast the future locations \cite{kothari2021human,kothari2021interpretable,liu2021social}.
These methods are generally effective when the trajectories are rather smooth, and future motions are similar to past ones.
However, as the past actions of people may not necessarily imply all of their future movements, trajectory-based methods react poorly to abrupt changes in pedestrian dynamics \cite{kothari2021human}.

We argue that predicting the stops and goes of pedestrians, i.e., the changes between the basic motion states of walking and standing still, can serve as a crucial component for more robust motion forecasting.
These transitions are one of the most common and essential aspects of human movement patterns.
However, they are highly non-linear in nature, and therefore usually hard to predict \cite{6856505, kooij2019context}.
Moreover, stops and goes are often involved in safety-critical traffic scenarios, such as a pedestrian waiting at the curbside and later walking for crossing \cite{jayaraman2020analysis,RAZALI2021103259}.
Failing to foresee such transitions can lead to catastrophic consequences. 

\begin{figure}[t]
    \centering
    \includegraphics[width=.94\linewidth]{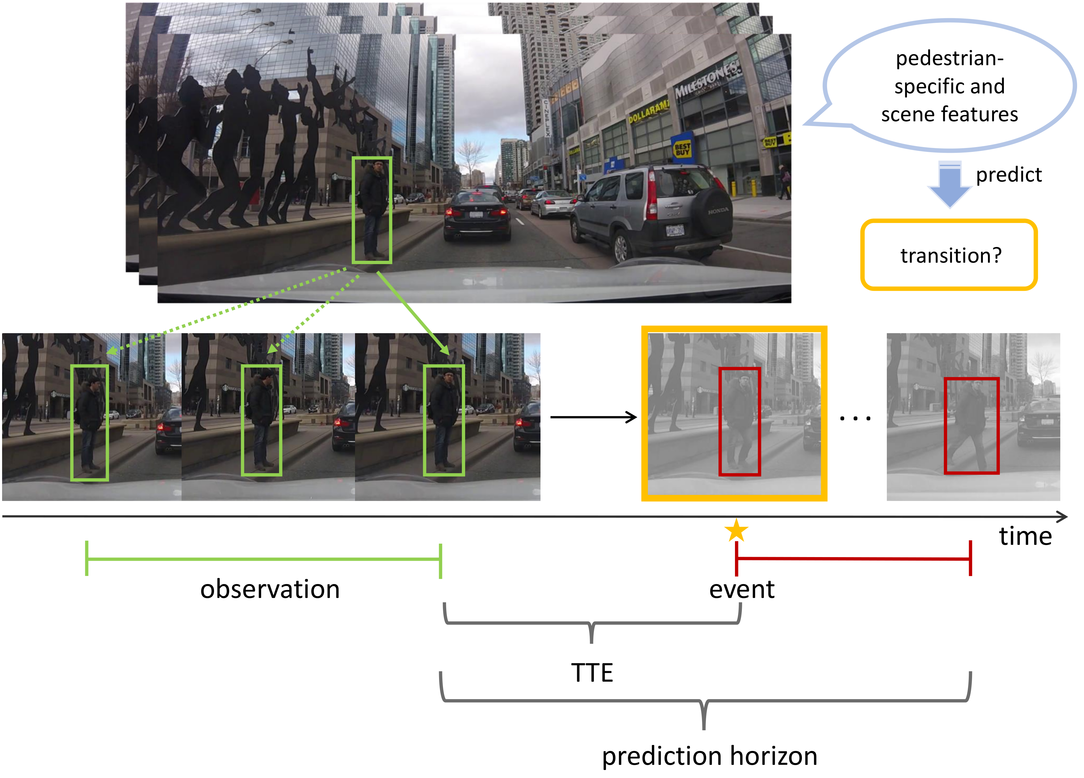}
    \caption{Illustration of a future transition forecasting. Given past observations, we predict whether a transition in the pedestrian's motion state will occur within the prediction horizon by reasoning over pedestrian-specific and scene features.
    Pedestrians are enclosed in green boxes when standing still, and in red boxes when walking.}
    \label{fig:task}
\end{figure}

In this paper, we introduce the task of pedestrian stop and go forecasting from the ego-centric view of a moving vehicle, as illustrated in \autoref{fig:task}.
In order to study it, we setup a benchmark with a new dataset and multiple approaches learned and evaluated on it.
For this, we first release TRANS, the first large-scale dataset for explicitly studying the stop and go behaviors of pedestrians in urban traffic.\footnote{Dataset available at \url{https://github.com/vita-epfl/pedestrian-transition-dataset}.}
It is based on several existing self-driving datasets annotated with pedestrians' walking motions, in order to have diversity in scenarios and environments.
Furthermore, we propose a hybrid model that fuses multi-modal inputs to capture both pedestrian-specific and contextual features in traffic scenes.\footnote{Code available at \url{https://github.com/vita-epfl/hybrid-feature-fusion}.}
Our model utilizes feed-forward and recurrent neural networks for spatial-temporal reasoning.
We also implement several baselines and analyze the impacts of various design choices.
In addition, detailed ablation experiments highlight the importance of contextual cues and temporal dynamics.

%%%%%%%%%%%%%%%%%%%%%%%%%%%%%%%%%%%%%%%%%%%%%%%%%%%%%%%%%%%%%%%%%%%%%%%%%%%%%%%%
\section{Related Work}

\subsection{Trajectory Forecasting}
Trajectory forecasting is an active area of research for modeling pedestrian behaviors. Many early works \cite{Helbing_1995, Berg_ICRA_2008, Burstedde2001SimulationOP} primarily focus on developing explicit models about pedestrian movements based on hand-crafted rules. Although they demonstrate some competitive results, these approaches impose strong priors and have limited capacity in capturing complex interactions. In recent years, data-driven methods that utilize neural networks to learn interactions have been shown to yield superior performances \cite{kothari2021human}. Alahi et al. introduce Social-LSTM \cite{Alahi_CVPR_2016} for predicting pedestrian trajectories in crowded spaces. Social-LSTM deploys a Long Short-Term Memory (LSTM) network for sequential modeling and integrates interactions of nearby pedestrians with a novel social pooling layer. Gupta et al. \cite{Gupta_CVPR_2018} use a Generative Adversarial Network (GAN) to learn and generate more socially acceptable trajectories. Attention mechanisms \cite{Ashish_NIPS_17} are also used to weight the influences of different neighbours on the person of interest \cite{sadeghian2019sophie, Yu2020SpatioTemporalGT}. The majority of trajectory prediction methods relies on top-down (bird's eye) views captured by stationary cameras. Malla et al. \cite{TITAN} explore using action priors from the view of a moving vehicle.

\subsection{Action Recognition and Early Prediction}
Before deep learning, methods using hand-crafted features, represented by Improved Dense Trajectories (IDTs) \cite{Heng_ICCV_2013}, were the state of the art of human activity recognition. Karpathy et al. achieve early action recognition at the frame level using Convolutional Neural Networks (CNNs) \cite{6909619}. Simonyan et al. \cite{Simonyan_NIPS_2014} introduce a two-stream network where a second CNN is added to learn temporal information in the videos based on optical flow streams. The great success of this last approach \cite{Simonyan_NIPS_2014} inspires following works to jointly model the spatial and temporal information in the videos \cite{zhu2020comprehensive}. Action prediction algorithms are inherently similar to recognition. Some commonly used methods include 3D convolution networks \cite{Ji_PAMI_2013, Tran_ICCV_2015}, recurrent networks \cite{Shi_2018_ECCV, abu2018will}, and more recently transformers \cite{girdhar2021anticipative}. Action recognition and early prediction methods have been applied to improve road safety, e.g., with accident estimation \cite{Yu_IROS_2019}, anticipating road crossing and pedestrians' intentions \cite{schneemann2016context, volz2016data, varytimidis2018action, saleh2019real, rasouli2019pedestrian, yao2021coupling}, and protecting vulnerable road users \cite{Kress_SSCI_2020, mordan2021detecting}.

\subsection{Stop/Go Detection and Prediction} 
Only a few previous works have explored the stop and go behaviors of road users in traffic. Keller et al. \cite{Keller_2013_TITS} detect stopping intentions of pedestrians moving toward the curbside using dense optical flow to predict future paths. Koehler et al. \cite{Sebastian_ITSM_2013} recognize pedestrian intentions to enter a street, to stop, and to bend in using Motion History Images (MHIs), HOG descriptors, and Support Vector Machines (SVMs). Quintero et al. \cite{Quintero_ITSC_2017, Quintero_TITS_2019} propose an approach to detect and predict pedestrian moving intentions utilizing a Hidden Markov Model (HMM) and body keypoints.
Kooij et al. \cite{6856505, kooij2019context} use Switching Linear Dynamics to integrate multiple motion modes into trajectory prediction. Apart from pedestrians, detecting the start intentions of cyclists is also investigated using 3D human pose \cite{Kress_ITSC_2019} or MHIs \cite{zernetsch2018early}. Despite obtaining solid results, these methods rely primarily on pedestrian-specific features such as position, velocity and body pose. The contextual and environmental cues, which can provide crucial information for long-term prediction, are largely ignored. In contrast, our work anticipates the stop and go behaviors by reasoning over both pedestrian-specific features and context information in the scene.

\begin{figure*}[t]
    \centering
    \includegraphics[width=.97\textwidth]{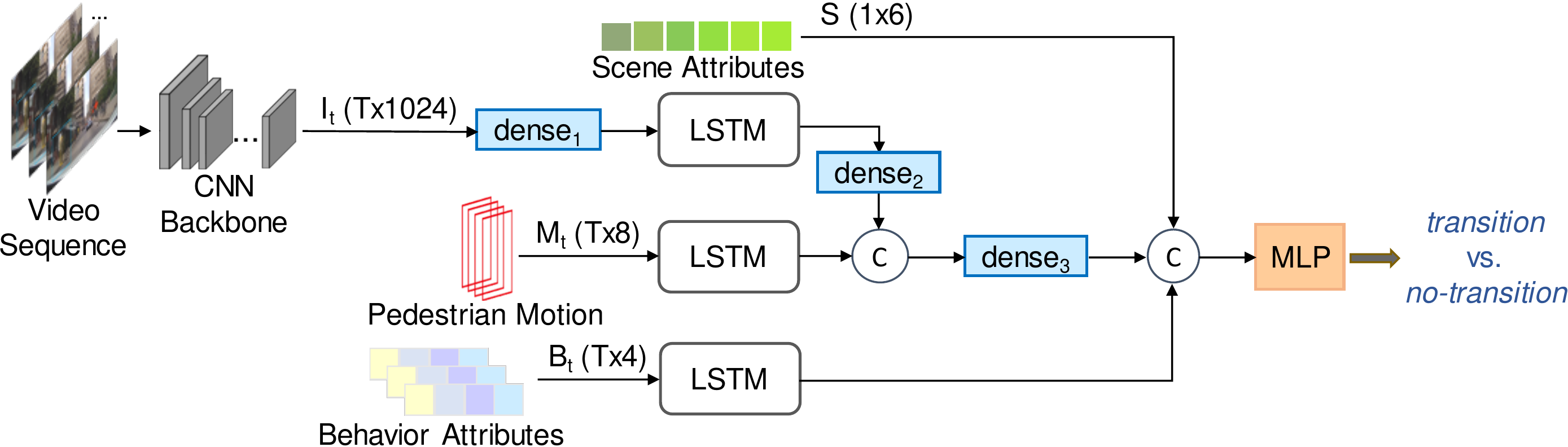}
    \caption{Overview of our proposed model. With a sequence of $T$ past observations, the model predicts whether a walking/standing pedestrian will stop/go within a time horizon $\lambda$. The past observations include video frames and high-level attributes, leading to four input modalities (\textit{Image}, \textit{Motion}, \textit{Behavior} and \textit{Scene} features) that are gradually fused together. A \textit{dense} block consists of a fully connected layer, a ReLU activation function, and dropout. Blocks \copyright{} represent concatenation. We test several CNN backbones for visual encoding and employ LSTMs for temporal processing.}
    \label{fig_PVIBS}
\end{figure*}

%%%%%%%%%%%%%%%%%%%%%%%%%%%%%%%%%%%%%%%%%%%%%%%%%%%%%%%%%%%%%%%%%%%%%%%%%%%%%%%%
\section{TRANS Dataset}

\begin{table}[t]
  \caption{Statistics of our TRANS dataset. \textit{Go}, \textit{Stop}, \textit{Stand}, \textit{Walk} indicate the number of unique pedestrians in corresponding categories. In brackets, we also count the number of events, i.e., stop and go transitions.}
  \label{tab:dataset_stats}
  \centering
  \begin{tabular}{lcccc}
    \toprule
    Dataset &  Go [events] & Stop [events]  & Stand  & Walk \\
    \midrule
    JAAD \cite{Rasouli_2017_ICCV_JAAD}   &     144 [145]   &  73 [77] & 65 & 416 \\
    \rowcolor{gray!15}
    PIE \cite{Rasouli_2019_ICCV_PIE}    &     397 [482]   & 528 [622] & 697 & 483 \\
    TITAN \cite{TITAN}  &     339 [381]   & 398 [439] & 1,077 & 6,233 \\
    \rowcolor{gray!15}
    TRANS  &     880 [1,008]  & 999 [1,138] & 1,839 & 7,132 \\
    \bottomrule
  \end{tabular}
\end{table}

To the best of our knowledge, there are no large-scale, real-world datasets currently available for studying the stops and goes of pedestrians in the context of autonomous driving. Hence, we build TRANS dataset to facilitate the research in this domain. It is built on top of several existing autonomous driving datasets annotated with walking behaviors of pedestrians (see \autoref{tab:dataset_stats}), so that it includes transition samples from diverse traffic scenarios with a unified interface.

\subsection{Benchmark Selection}
We augment three existing self-driving datasets, namely JAAD \cite{Rasouli_2017_ICCV_JAAD}, PIE \cite{Rasouli_2019_ICCV_PIE} and TITAN \cite{TITAN} for building the benchmark.  The three are closely related to our task in the sense that they all provide RGB videos captured from an uncalibrated monocular camera on a moving platform, together with localization and walking annotations for pedestrians. They also provide train, validation, and test splits by video clips.

\subsubsection{Joint Attention for Autonomous Driving (JAAD)}
it aims at exploring the road users' behaviors in pedestrian crossing settings \cite{Rasouli_2017_ICCV_JAAD}. It consists of 346 short video snippets recorded at 30fps with a dashboard camera under various weather and lighting conditions. JAAD provides 2D bounding boxes for all pedestrians, among which 654 are around potential crossing events and are annotated with walking labels.

\subsubsection{Pedestrian Intention Estimation (PIE)}
it is designed for the task of recognizing pedestrians' intentions of crossing the roads \cite{Rasouli_2019_ICCV_PIE}. It is sourced from 6 hours of continuous daytime driving recorded at 30fps by a monocular camera in North America. Labels that indicate motions
are available for 1,842 pedestrians close to the road that may potentially interact with the driver.

\subsubsection{Trajectory Inference using Targeted Action priors Network (TITAN)}
it is a dataset recently introduced for trajectory forecasting and multi-level action recognition \cite{TITAN}. 10 hours of driving video are recorded at 60fps in densely populated central Tokyo. To construct the final dataset, 700 short clips are extracted and annotated at a sampling frequency of 10Hz. TITAN contains 8,592 unique pedestrians with multiple action labels organized hierarchically by contextual complexity. The labels for primitive actions such as walking, standing still and running are mutually exclusive.

Overall, JAAD contains selected short clips centered on road crossings, while PIE focuses on all potential crossings in a more general way. TITAN is even more generic with numerous annotations on pedestrians not interacting with the traffic. TRANS therefore offers an increasing level of difficulty through the various datasets, by varying the relevance of objective cues related to traffic.

\subsection{Annotation pipeline}
To simplify the annotation process, we rely on the original annotations of walking motions in each dataset to detect transitions. We count a state change from \textit{standing} to \textit{walking} as a \textit{Go} candidate, and the opposite as a \textit{Stop} one. In TITAN, the activities like \textit{running} are also viewed as \textit{walking}. We refer to the state before the transition as \textit{pre-state} and the state after as \textit{post-state}. To reduce possible errors by inaccurate labeling and obtain more meaningful samples, we only consider a transition candidate to be valid if the durations of its \textit{pre-state} and \textit{post-state} both last longer than 0.5 seconds. 

All unique pedestrians in the original datasets can be categorized into \textit{Walk}, \textit{Stand}, \textit{Stop} and \textit{Go}. \textit{Walk} and \textit{Stand} pedestrians show no transitions throughout the video, whereas \textit{Stop} and \textit{Go} pedestrians exhibit the corresponding transitions. The classes of \textit{Stop} and \textit{Go} are not mutually exclusive since a pedestrian can perform both stops and goes during the same observation. The statistics of datasets are presented in \autoref{tab:dataset_stats}.

After inspecting the detected transitions, we find the majority of stops and goes in JAAD and PIE to be closely related to road crossings. Compared to JAAD, PIE contains more non-crossing transitions and edge cases. However, causes for transitions in TITAN  are diverse and often ambiguous. 

\subsection{Problem Formulation}
We formulate pedestrian stop and go forecasting as a binary classification problem, and illustrate it in \autoref{fig:task}. Given a sequence of past observations of length $T$, the objective is to determine whether a given walking/standing pedestrian will stop/go within a time horizon $\lambda$. The observations include video frames with additional pedestrian and scene attributes. We assume the pedestrian's current bounding box and motion state (standing/walking) are known, and the model output is a binary prediction of \textit{transition} vs. \textit{no-transition}. The stops and goes are evaluated as two separate tasks.

%%%%%%%%%%%%%%%%%%%%%%%%%%%%%%%%%%%%%%%%%%%%%%%%%%%%%%%%%%%%%%%%%%%%%%%%%%%%%%%%
\section{Hybrid Feature Fusion}

Diverse social and environmental factors have been shown to impact pedestrian motions and decision-making in urban traffic \cite{kotseruba2020joint}. We propose a hybrid model for pedestrian stop and go forecasting that encodes pedestrian-specific features jointly with dynamics and contextual information. Our model relies on both feed-forward and recurrent structures to process multi-modal inputs. The detailed model design is shown in \autoref{fig_PVIBS} and is described in the following.

\begin{figure}[t]
    \centering
    \includegraphics[width=\linewidth]{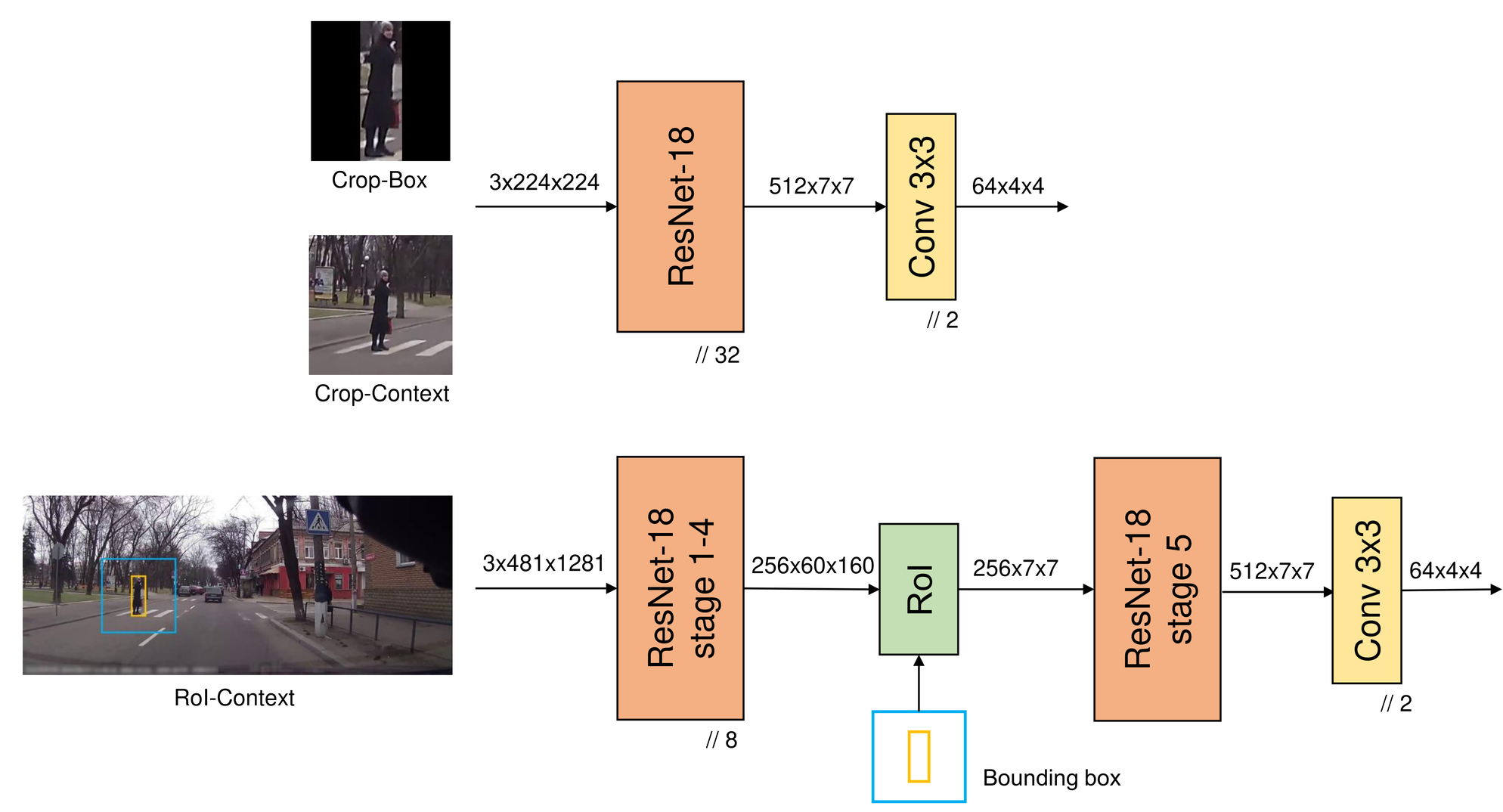}
    \caption{CNN backbones for visual encoding. For \textit{Crop-Box} and \textit{Crop-context} (top), the inputs are RGB image crops, either by original or enlarged pedestrian bounding boxes, scaled to size $224 \times 224$. The feature extractor is a ResNet-18 backbone followed by a $3 \times 3$ convolution. For \textit{RoI-Context} (bottom), the input is the whole image of size $481 \times 1281$, and a RoI-alignment layer, using the enlarged pedestrian bounding box (in blue) as the region proposal, is inserted between the ResNet's fourth and fifth stages.}
    \label{fig:CNN_backbone}
\end{figure}

\textbf{Visual Encoding.}
We process each image with a Convolutional Neural Network (CNN) to extract information about pedestrians and the contexts around them.
We implement several CNN backbones that include different levels of context, as shown in \autoref{fig:CNN_backbone}, to analyze the impact of the context.

First, we compare between no or local contexts around pedestrians.
Without context, we crop every image at the pedestrian bounding box and pad it with zeros to make it square.
To get local context, we extract a square image patch around the pedestrian by scaling up the corresponding bounding box by 2, then matching the scaled box's width with its height.
For both methods, the patches are fed to a ResNet-18 \cite{Kaiming_CVPR_2016} backbone for feature extraction.
We refer to the former as \textit{Crop-Box} and to the later as \textit{Crop-Context}.

We also implement a CNN backbone to extract the visual features from whole images instead of patches.
We modify the previous ResNet-18 backbone by inserting a RoI-alignment layer \cite{Kaiming_ICCV_2017} between the fourth and fifth stages to capture broad contexts in the original images.
For each frame, the RoI is defined by the pedestrian's bounding box enlarged following the same procedure as for \textit{Crop-context}.
To better preserve the resolution, we also remove the max pooling in the ResNet's first stage and modify the fifth stage so that the first convolution operates on a $7\times7$ feature map with a stride of 1.
We refer to this design as \textit{RoI-Context}.

After the ResNet backbone, we add a $3\times3$ convolution to reduce the dimension, and flatten the output to get the visual feature $I_t$ for the frame $t$.

\vspace{2em}
\textbf{Motion Encoding.}
We encode the motions of pedestrians by collecting their positions and velocities. A 4D vector $P_t$ represents a pedestrian's position at each time step $t$:
\begin{equation}
    P_t = (x_t, y_t, w_t, h_t),
\end{equation}
where ($x_t$, $y_t$) are the x-y coordinates of the center of the pedestrian's bounding box, and $w_t$, $h_t$ are the box's width and height.
The velocity $V_t$ at time step $t$ is then defined as the change in position from the previous frame $t-1$, with a time difference of $\Delta t$ between both frames:
\begin{equation}
\begin{aligned}
    V_t &= (\Delta x_t, \Delta y_t, \Delta w_t, \Delta h_t) \\
        &= \frac{P_t - P_{t-1}}{\Delta t}.
\end{aligned}
\end{equation}
The position and velocity vectors together form our motion feature vector $M_t = (P_t, V_t)$ at each time step $t$.

\textbf{Behavior Encoding.}
Specific body language may reflect pedestrians' will to communicate with the drivers or imply future motions. We use binary attributes to indicate three non-verbal behaviors: looking, nodding, and hand gestures. Additionally, we include the walking state, making the behavioral feature $B_t$ a 4D binary vector at each frame $t$:
\begin{equation}
    B_t = (b_{walk}^t, b_{look}^t, b_{nod}^t, b_{hand}^t).
\end{equation}
These behavioral attributes are only available in JAAD and PIE.
Although hand gestures can cover a wide range of meanings, we only use a binary attribute as done in these datasets. Having further distinctions would likely lead to better and more confident predictions when gestures are recognized.

\textbf{Scene Encoding.}
JAAD and PIE provide six high-level semantic attributes that form a coarse, static representation $S$ of the traffic scene:
\begin{equation}
    S = (s_{tl}, s_{in}, s_{de}, s_{si}, s_{td}, s_{md}),
\end{equation}
where $s_{tl}$ denotes the number of traffic lanes, binary variables $s_{in}$, $s_{de}$, $s_{si}$ indicate whether the scene is at an intersection, and whether this intersection is designated (with a zebra crossing or a traffic signal) or signalized, and $s_{td}, s_{md}$ show the traffic direction (one-way or two-way) and pedestrian motion direction (lateral or longitudinal).
These semantic attributes are not available in TITAN.

\textbf{Temporal Processing.}
Recurrent Neural Networks (RNNs) have proven to be highly successful in sequential learning \cite{lipton2015critical}. Long Short-Term Memory (LSTM) networks, as variations of RNNs, address the problem of vanishing gradients and long-term dependency in modeling long sequences \cite{NIPS2014_a14ac55a}. We propagate visual, motion and behavior features through separate LSTMs for temporal processing, and obtain the hidden states at the final time step.

\textbf{Hybrid Fusion.}
Inspired by \cite{rasouli2020multimodal}, we adopt a hybrid fusion strategy where inputs and intermediate features are individually processed and then concatenated in a gradual fashion, as observed in \autoref{fig_PVIBS}. We use dense layers to process features and reduce dimensions. The joint representation of all input modalities is then fed to a Multi-Layer Perceptron (MLP) to yield the final predictions.

%%%%%%%%%%%%%%%%%%%%%%%%%%%%%%%%%%%%%%%%%%%%%%%%%%%%%%%%%%%%%%%%%%%%%%%%%%%%%%%%
\section{Experiments}

\subsection{Data Preparation}
We conduct all experiments on our new TRANS dataset.
Examples consist of video sequences of past observations of length $T$. Each frame possesses a time-to-event (TTE) tag indicating the time gap between this frame and the subsequent stop or go transition. The label of a video sequence (\textit{transition} or \textit{no-transition}) is determined by the TTE tag of its last frame. As crossing-related cases are arguably the most critical for stop and go forecasting, we set the prediction time horizon $\lambda$ to be 2 seconds, which is the minimum time within which pedestrians make crossing decisions \cite{Schmidt2009PedestriansAT}.  We choose a relatively large sampling rate of 5fps in the hope of reducing overfitting and speeding up training. Since the pedestrians far away from the ego vehicle are of less interest, we remove the instances whose widths of the pedestrian bounding boxes in the last frames are smaller than 24 pixels.
It is important to note that we use ground-truth boxes and attributes as inputs here. However, these would first need to be predicted by another model in practice, and noise in the predictions would likely negatively affect the final results.

\subsection{Models}
We compare the performance of the proposed model to a series of baselines, grouped into \textit{Static}, \textit{Video} and \textit{Hybrid}.

\textbf{Static.}
Static baselines take single image frames as input, and directly output the classification results. We combine the visual encoders mentioned previously (\textit{Crop-Box} (CB), \textit{Crop-Context} (CC), \textit{RoI-Context} (RC)) with a fully connected layer to yield the predictions (i.e., without using LSTM).

\textbf{Video.}
Video baselines extract the visual features from video sequences and use LSTMs for temporal encoding.
They use the same visual encodings as \textit{Static} baselines.

\textbf{Hybrid.}
Hybrid baselines use high-level attributes as input in addition to videos.
The first model utilizes two input modalities available in all three datasets: \textit{Images} (I) using a \textit{RoI-Context} (RC) encoder, and pedestrian \textit{Motion} (M). Its architecture is consistent to the design in \autoref{fig_PVIBS}. Our full model also uses \textit{Behavioral} (B) and \textit{Scene} (S) attributes.

\subsection{Implementation Details}
All backbones for image processing are ResNet-18 \cite{Kaiming_CVPR_2016}.
For recurrent networks, we use vanilla LSTMs with \textit{tanh} activation, and the sizes of hidden states for encoding \textit{Images}, \textit{Motion}, and \textit{Behavior} features are set to 256, 64, and 16 respectively. We set the sizes of the three embedding dense layers to 256, 128 and 64 in order. The MLP for final prediction consists of three fully connected layers of size 86, 86 and 1. We apply dropout with a rate of 0.2 in dense layers for regularization.
If not specified, we set the input observation sequence length to $T=5$.

Training consists of two stages.
First, the weights of the visual encoder CNN are obtained by training the corresponding \textit{Static} baseline \textit{RoI-Context} for the same classification task. At this training stage, we augment the images with random horizontal flipping, cropping out of the top third, resizing to $481 \times 1281$, random color jittering, and random grayscale conversion.
We then freeze the ResNet backbone and train other parts of the model. The models are trained with Adam \cite{Kingma2015AdamAM} optimizer with a batch size of 8, learning rate of $1 \cdot 10^{-4}$, and weight decay of $1 \cdot 10^{-5}$. We use a binary cross entropy (BCE) loss function.
We employ early stopping during training and the number of epochs for convergence varies for each dataset. To compensate for the data imbalance, we randomly sample the over-represented class during training.

\subsection{Evaluation Metrics}
We use Average Precision (AP) to evaluate the models' performances. As each dataset is unbalanced toward \textit{no-transition}, we calculate the AP on a balanced test set where negative instances are randomly sampled. To reduce the variance introduced by sampling, we conduct 10 randomized trials and report the averaged results for each model.

\subsection{Quantitative Results}

\begin{table}[t]
  \caption{Results in Average Precision (AP, \%) on TRANS dataset. Modalities are \textit{Images} (I), \textit{Motion} (M), \textit{Behavior} (B) and \textit{Scene} (S). Visual contexts are \textit{Crop-Box} (CB), \textit{Crop-Context} (CC) and \textit{RoI-Context} (RC).}
  \label{tab:AP_gO_stop}
  \centering
  \addtolength{\tabcolsep}{-1.6pt}
  \addtolength{\cmidrulekern}{-1.6pt}
  \begin{tabular}{llcccccc}
    \toprule
    & & \multicolumn{3}{c}{Go} &      \multicolumn{3}{c}{Stop}\\
    \cmidrule(lr){3-5} \cmidrule(lr){6-8}
    Model & Modalities & JAAD & PIE & TITAN  & JAAD & PIE & TITAN \\
    \midrule
    & I (CB)   &   54.3   &  52.0    &  56.2 &    52.5 & 53.1    &  56.4   \\
    & I (CC)  & 70.4     & 59.1     & 61.4 & 57.3       &  61.1   & 60.3\\
    \multirow{-3}{*}{Static} & I (RC)  &  73.3    &  61.2    &   60.9  & 58.7  &   62.5    &   59.1 \\
    \addlinespace
    \rowcolor{gray!15}
    & I (CB) & 60.6 & 56.4 & 58.6 &  57.2  & 59.4 &  58.7\\
    \rowcolor{gray!15}
    & I (CC)  & 73.6 & 61.8 & 63.2  & 61.4 & 63.3 & 61.5\\
    \rowcolor{gray!15}
    \multirow{-3}{*}{Video} & I (RC)  & 76.4 & 64.7 & 62.9  & 62.9 & 64.2 & 61.7\\
    \addlinespace
    & IM (RC) & 80.6 & 66.5 & \textbf{65.1} & 64.7 & 64.9  & \textbf{63.6}\\
    \multirow{-2}{*}{Hybrid} & IMBS (RC) & \textbf{85.9}  & \textbf{70.2} & -- &  \textbf{67.8} & \textbf{65.4} & --\\
    \bottomrule
  \end{tabular}
  \addtolength{\cmidrulekern}{1.6pt}
  \addtolength{\tabcolsep}{1.6pt}
\end{table}

\autoref{tab:AP_gO_stop} summarizes the stop and go forecasting results with AP metric. Our full model achieves the best performance on JAAD and PIE. In particular, for go forecasting, compared to the best \textit{Video} baseline, the full model improves AP by 9.5 points on JAAD and 5.5 points on PIE. The improvements are expected as the high-level attributes of the pedestrians and traffic scenes have strong correlations with crossing, which is the primary cause for go transitions in JAAD and PIE. Interestingly, the improvements in stop forecasting are less noticeable, which may suggest the stops of pedestrians are less correlated with the behavioral and semantic attributes.

On the three datasets, the \textit{Hybrid} model that fuses pedestrian motion features with visual cues outperforms all \textit{Video} baselines. Comparing the results of \textit{Static} baselines, we see that adding visual representation of the context improves AP by a large margin. Using global context yields better results than local context on JAAD and PIE but is inferior on TITAN. The notable improvements in AP from \textit{Static} baselines to \textit{Video} ones demonstrate the benefits of using sequential models for temporal reasoning.

On all the datasets, the models' performances for go prediction are generally better than for stop. This could be caused by the fact that, in a typical crossing scenario, go predictions may benefit more from specific high-level attributes of the pedestrian behaviors and the scene, such as body languages and the existence of designated crossing, while the social cues for stops can sometimes be ambiguous or not present in our attributes. As a result, we can see a wide gap between the stop and go predictions on JAAD, which is mainly about crossing scenarios. On the other hand, the performance gap shrinks on PIE, which contains more non-crossing cases, and the results of these two tasks are close on TITAN, which focuses less on crossing.

\subsection{Qualitative Analysis}
\begin{figure*}[t]
    \centering
    \subfloat[Go transitions]{%
        \includegraphics[width=.82\linewidth]{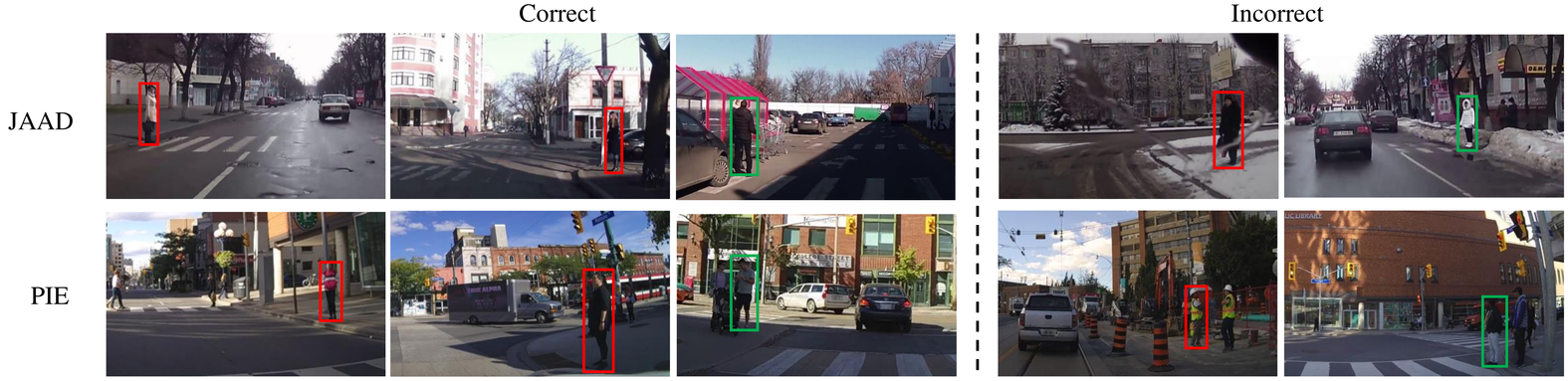}}
    \hfill \\
    \subfloat[Stop transitions]{%
        \includegraphics[width=.82\linewidth]{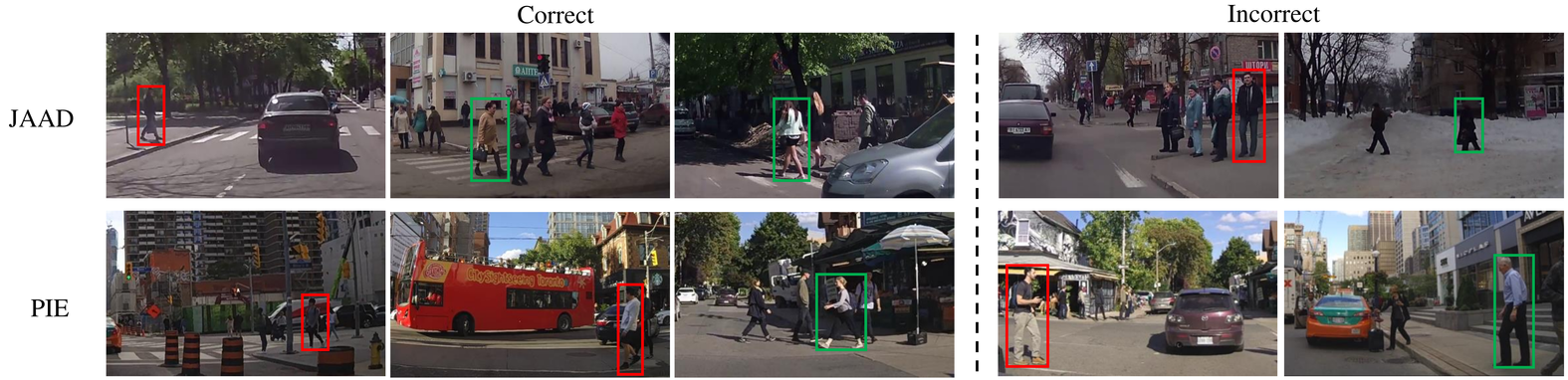}}
    \caption{Qualitative results of our full proposed \textit{Hybrid} model on JAAD \cite{Rasouli_2017_ICCV_JAAD} and PIE \cite{Rasouli_2019_ICCV_PIE} datasets. The predictions for future transitions and non-transitions are indicated by red and green boxes respectively. The results are grouped by \textit{Go} (top) and \textit{Stop} (bottom) forecasting with correct predictions on the left and incorrect ones on the right.}
    \label{fig:qualitative} 
\end{figure*}

\autoref{fig:qualitative} displays qualitative example predictions of our proposed full \textit{Hybrid} model on JAAD and PIE datasets. We can see that the stops and goes of pedestrians at crossroads remain challenging to predict, partially due to the lack of ego vehicle speed and states of the traffic lights. In addition, sudden changes of moving direction, weather conditions (e.g., rainy, snowy), and irregular cases like construction workers can also negatively impact the predictions.

\subsection{Ablation Study}
We have discussed the importance of multi-modal fusion and temporal processing.
We now conduct ablation experiments to further investigate the contributions of different input modalities and the impact of the length $T$ of the observations.

\subsubsection{Choice of Modalities}
We assess the contributions of individual features in \autoref{tab:abl_features} by experimenting with different combinations of input modalities for \textit{Hybrid} models.
Adding \textit{Behavior} (B) and \textit{Scene} (S) attributes to \textit{Motion} (M) information boosts the performance.
The improvements on AP metric are particularly important for go forecasting, by up to 23.2 points on JAAD and 7.5 points on PIE.
Adding \textit{Image} (I) features further enhances the models' performances.
It should be noted that when combined with high-level attributes, the performance gaps between local (\textit{Crop-Context}) and global (\textit{RoI-Context}) visual contexts are narrow.

\begin{table}[t]
  \caption{Ablation study in Average Precision (AP, \%) on the choice of modalities for \textit{Hybrid} models. Modalities are \textit{Images} (I), \textit{Motion} (M), \textit{Behavior} (B) and \textit{Scene} (S). Visual contexts are \textit{Crop-Context} (CC) and \textit{RoI-Context} (RC).}
  \label{tab:abl_features}
  \centering
  \begin{tabular}{lcccc}
    \toprule
    & \multicolumn{2}{c}{Go} & \multicolumn{2}{c}{Stop} \\
    \cmidrule(lr){2-3} \cmidrule(lr){4-5}
    Modalities & JAAD & PIE & JAAD & PIE\\
    \midrule
    S & 74.2 & 55.1 & 53.3 & 54.2\\
    \rowcolor{gray!15}
    M & 61.5 & 59.8 & 59.4 & 60.6\\
    I (CC) & 73.6 & 61.8 & 61.4 & 63.3\\
    \rowcolor{gray!15}
    I (RC) & 76.4 & 64.7 & 62.9 & 64.2\\
    IM (CC) & 78.4 & 65.1 & 63.4 & 63.5\\
    \rowcolor{gray!15}
    IM (RC) & 80.6 & 66.5 & 64.7 & 64.9\\
    MBS & 84.7 & 67.3 & 62.5 & 64.7\\
    \rowcolor{gray!15}
    IMBS (CC) & 85.2 & 69.5 & 67.2 & \textbf{65.7}\\
    IMBS (RC) & \textbf{85.9} & \textbf{70.2} & \textbf{67.8} & 65.4 \\
    \bottomrule
  \end{tabular}
\end{table}

\subsubsection{Impact of the Length of Observation Sequences}
In \autoref{tab:abla_temporal}, we study how the prediction performance evolves as we change the length $T$ of observation sequences. Overall, the predictions improve when more observations come in. However, the performance improvements reach saturation at some point, as evidenced by the stagnation or decrease in AP when we extend the input length from 10 to 15 frames. This behavior is expected as earlier frames should be less correlated with the later transitions.

\begin{table}[t]
  \caption{Ablation study in Average Precision (AP, \%) on the length $T$ of observation sequences. Modalities are \textit{Images} (I), \textit{Motion} (M), \textit{Behavior} (B) and \textit{Scene} (S). Visual contexts are \textit{RoI-Context} (RC).}
  \label{tab:abla_temporal}
  \centering
  \begin{tabular}{lccccc}
    \toprule
    &  & \multicolumn{2}{c}{Go} & \multicolumn{2}{c}{Stop} \\
    \cmidrule(lr){3-4} \cmidrule(lr){5-6}
    Model & T  &  JAAD & PIE & JAAD & PIE \\
    \midrule
    & 1 & 72.5 & 61.8  & 55.8 & 61.3 \\
    & 5 & 76.4 & 64.7  & 62.9 & 64.2 \\
    & 10 & \textbf{76.9} & 65.8  & \textbf{63.4} & \textbf{66.1} \\
    \multirow{-4}{*}{Video -- I (RC)}  & 15 & 74.8 & \textbf{66.2}  & 60.7 & 65.7 \\
    \addlinespace
    \rowcolor{gray!15}
    & 1 & 73.6 & 62.6  & 59.7 & 62.0 \\
    \rowcolor{gray!15}
    & 5 & 85.9 & 70.2 & 67.8 & 65.4 \\
    \rowcolor{gray!15}
    & 10 & \textbf{86.7} & 71.5  & \textbf{68.4} & \textbf{67.9} \\
    \rowcolor{gray!15}
    \multirow{-4}{*}{Hybrid -- IMBS (RC)}  & 15 & 85.1 & \textbf{71.9}  & 64.3 & 67.2 \\
    \bottomrule
  \end{tabular}
\end{table}

%%%%%%%%%%%%%%%%%%%%%%%%%%%%%%%%%%%%%%%%%%%%%%%%%%%%%%%%%%%%%%%%%%%%%%%%%%%%%%%%
\section{Conclusions}

In this paper, we have introduced the problem of pedestrian stop and go forecasting.
Accurately predicting these highly non-linear transitions is crucial to understand pedestrians' trajectories and guarantee their safety.
To foster research on this problem, we have also set up a new benchmark for the community.
For this, we have released TRANS, the first large-scale dataset for pedestrian stop and go forecasting from a vehicle perspective.
It is based on several existing datasets in order to contain various scenarios and environments.
We have then introduced a new deep learning model leveraging video sequences and high-level attributes about pedestrians and contextual scenes through a hybrid feature fusion, and have evaluated it, along with multiple baselines, on TRANS.
We have finally performed extensive experiments to analyse the impact of all the components and design choices.

\section{Acknowledgments}
We would like to thank Valeo and EPFL for funding our work, and the reviewers for their helpful comments.

%\addtolength{\textheight}{-10.5cm}   % This command serves to balance the column lengths
                                  % on the last page of the document manually. It shortens
                                  % the textheight of the last page by a suitable amount.
                                  % This command does not take effect until the next page
                                  % so it should come on the page before the last. Make
                                  % sure that you do not shorten the textheight too much.

%%%%%%%%%%%%%%%%%%%%%%%%%%%%%%%%%%%%%%%%%%%%%%%%%%%%%%%%%%%%%%%%%%%%%%%%%%%%%%%%
\bibliographystyle{IEEEtran}
\bibliography{biblio}

\end{document}